# Research on Image Super-Resolution Reconstruction Mechanism based on Convolutional Neural Network

Hao Yan *

College of Engineering and Computer Science, Syracuse University, Syracuse, NY, USA, hyan17@syr.edu

Zixiang Wang

College of Engineering and Computer Science, Syracuse University, Syracuse, NY, USA, zwang161@syr.edu

Zhengjia Xu

SATM, Cranfield University, Cranfield, UK, Billy.Xu@cranfield.ac.uk

Zhuoyue Wang

Department of Electrical Engineering and Computer Sciences, University of California, Berkeley, Berkeley, CA

zhuoyue_wang@berkeley.edu

Zhizhong Wu

Independent Researcher, Mountain View, CA, USA, ecthelion.w@gmail.com

Ranran Lyu

Amazon Sagemaker GoundTruth, Amazon Web Services, San Jose, CA, USA, rranlyu@gmail.com

**Abstract**

Super-resolution reconstruction techniques entail the utilization of software algorithms to transform one or more sets of low-resolution images captured from the same scene into high-resolution images. Nevertheless, the extraction of image features and nonlinear mapping methods in the reconstruction process remain challenging for existing algorithms. These issues result in the network architecture being unable to effectively utilize the diverse range of information at different levels. The loss of high-frequency details is significant, and the final reconstructed image features are overly smooth, with a lack of fine texture details. This negatively impacts the subjective visual quality of the image. The objective is to recover high-quality, high-resolution images from low-resolution images. In this work, an enhanced deep convolutional neural network model is employed, comprising multiple convolutional layers, each of which is configured with specific filters and activation functions to effectively capture the diverse features of the image. Furthermore, a residual learning strategy is employed to accelerate training and enhance the convergence of the network, while sub-pixel convolutional layers are utilized to refine the high-frequency details and textures of the image. The experimental analysis demonstrates the superior performance of the proposed model on multiple public datasets when compared with the traditional bicubic interpolation method and several other resolution methods.

## 1 INTRODUCTION

In consequence of the accelerated growth of the social economy in the contemporary era, the public's appetite for the conveyance and reception of information has increased markedly. Since the beginning of the 21st century, there has been a notable advancement in internet information technology, particularly in the context of the rapid growth of digital technology. Artificial intelligence has emerged as a pivotal force in people's daily lives and production activities. The scientific research achievements in the field of artificial intelligence, such as machine learning, natural language processing, computer vision, and image recognition, are gradually being transformed from theoretical exploration to important tools that are being widely used in reality [1]. In practical applications, the clarity of images is of paramount importance, as they are an intuitive and widely used medium for conveying information. The resolution of the image directly affects the quality of the information conveyed.

The resolution of an image, defined as the number of pixels per inch, is a critical indicator of the image's visual richness and complexity. An increase in the number of pixels results in an enhanced texture, detail, and clarity of the image [2]. However, the quality of images is frequently compromised by the adverse conditions of the acquisition environment, including optical contamination, motion blur and the limitations of hardware devices, such as inaccurate focus. Furthermore, during the transmission of images, the necessity for processing and compression may result in the introduction of noise, which subsequently impacts the visual effects and accurate transmission of information [3].

To address these issues, contemporary image processing methodologies, particularly those pertaining to image super-resolution reconstruction and artificial intelligence, offer a robust solution. These technologies facilitate the recovery of high-resolution details from low-resolution images, thereby markedly enhancing image quality and the visual experience [4]. The employment of deep learning models, such as generative adversarial networks (GANs) and convolutional neural networks (CNNs), enables the advancement of these techniques to not only enhance the resolution of images but also to effectively mitigate image loss and optimize the overall quality and visual effect of images without the necessity of additional sensors or hardware devices. This is of great significance in ensuring the accuracy of information and enhancing the user experience [5].

The primary function of image super-resolution reconstruction technology is to enhance the high-frequency information present in an image. This is achieved by processing one or more low-resolution image data sets, while simultaneously removing any unwanted high-frequency information, such as additive noise. This results in a reconstructed image with a greater number of pixels per unit size, thereby enhancing the overall quality of the image and significantly improving the visual effect through the introduction of richer texture detail. Single-image super-resolution represents a particularly complex challenge within the field of image super-resolution [6]. This is due to the fact that it involves the recovery of lost high-resolution information from limited low-resolution data, a process that is inherently unstable. Furthermore, a specific low-resolution input may correspond to a multitude of high-resolution outputs, and the relationship between the low and high resolutions is typically not evident. Additionally, the process of establishing a high-dimensional mapping is often time-consuming and inefficient [7].

Image super-resolution represents a pivotal computer vision technique, whereby high-resolution (HR) images are reconstructed from low-resolution images. This technology has demonstrated considerable potential and practical value in a number of application areas, including satellite imaging, medical imaging, video enhancement and security surveillance [8]. The rapid development of artificial intelligence and machine learning technologies has led to the emergence of image super-resolution reconstruction methods based on deep learning, particularly those based on convolutional neural networks, which have become a prominent area of research in this field. Convolutional neural networks are particularly adept at image processing tasks due to their robust feature extraction capabilities. Convolutional neural networks are capable of automatically learning and extracting high-level features from images, which is particularly advantageous when addressing image super-resolution problems [9]. The capacity of convolutional neural networks to process multi-layered information, from low-level textures to high-level semantics, is a key factor in their ability to reconstruct high-quality detail and textures.

The traditional interpolation-based methods, such as bilinear and bicubic interpolation, are relatively straightforward but often yield only limited improvements and are unable to recover high-frequency details. In contrast, deep learning methods are capable of learning more complex and detailed image reconstruction rules by training a large number of low-high-resolution image pairs, thereby significantly improving the quality of the reconstructed images. Furthermore, a number of innovative convolutional neural network (CNN) models proposed in recent years, including generative adversarial networks (GANs) and residual learning networks, have significantly advanced the state of the art in image super-resolution technology.

## 2 RELATED WORK

The interpolation-based super-resolution reconstruction algorithm is mainly focused on creating high-resolution images from the low-resolution data of images. Such algorithms, including the most commonly used nearest neighbor interpolation, bilinear interpolation, and bicubic interpolation, essentially increase the number of pixels in an image, i.e., the number of pixels per unit size in the internal region of the image. The reason why these traditional methods are widely used in the field of image super-resolution is mainly due to their fast computational speed and simple algorithms, which can quickly complete the super-resolution reconstruction of images with only a few algorithm resources. These interpolation-based methods (INTER) [10] are particularly important when working with a single low-resolution image in a real-world working environment, as they are able to process the image data in real-time to meet the needs of the production line or immediate feedback system. However, while these interpolation techniques are efficient at solving problems and have a limited resource footprint, they are non-adaptive methods, which means that they cannot be optimized for the specific content of the image. Therefore, when reconstructing images at super-resolution, these methods may cause some quality problems, such as aliasing, ghosting, blurring, and glitches, which may affect the overall quality of the reconstructed image.

These problems often arise because interpolation algorithms only mathematically smooth the existing pixels and are unable to recover or reconstruct the high-frequency details that are lost in the image. To overcome these limitations, researchers have begun to explore more advanced methods, such as learning-based super-resolution techniques, including the use of deep learning and machine learning models, which can learn from large amounts of data how to reconstruct image details more efficiently. By understanding the content and structural features of the image, these advanced methods can provide more accurate reconstruction results, significantly improving the quality of the final image.

Yeung et al. [11] employed the manifold learning (ML) method for super-resolution reconstruction, operating under the assumption that in the feature space, the low-resolution image data and the corresponding high-resolution images exhibit consistency in their local manifold features. By employing domain embedding techniques, this method is capable of aligning the structural data of the low-dimensional space with that of the high-dimensional space. During the training phase, the researchers obtain the fitting coefficients and map the low-resolution image data to the structural representation of the high-dimensional image data. In order to minimize the reconstruction error, the least squares method and local linear embedding techniques are employed to calculate the weighted parameters. Ultimately, the structural data of these high-dimensional spaces are combined with the weighting coefficients to generate reconstructed image data with minimal error. Furthermore, in order to reduce the complexity of the expression coefficient, the researchers employed the corresponding high-resolution images to reconstruct the low-resolution images, thereby further optimizing the reconstruction process.

In contrast, Wright and his team [12] have developed a method that postulates the sharing of sparsity coefficients (SC) between low and high resolution images, based on their study of compressive sensing. In this approach, the pertinent image feature parameters are generated by transforming the image data into a sparse domain, utilizing a dictionary of low-resolution and high-resolution image data. In conclusion, the aforementioned parameters are employed in order to generate a super-resolution reconstructed image. The advantage of this method is that it circumvents the necessity to artificially determine the number of adjacent images. However, accurately identifying noisy data represents a significant challenge, and the method is also susceptible to the limitation of the number of dictionaries, which may result in reconstructed images with insufficiently accurate edge details. The method based on image self-similarity offers an alternative perspective, proposing that the local details, texture content, and structural scale of low-resolution images can manifest repeatedly in the context of high-dimensional data. Appropriate down-sampling methods enable this strategy to perform complex scale transformations on the input image data, thereby generating multi-resolution image training samples. This method exploits the repetitive texture pattern inherent in the image and employs advanced data processing techniques to enhance the effect and efficiency of super-resolution reconstruction, thus demonstrating its potential in processing image details and maintaining structural coherence.

## 3 METHODOLOGIES

In this section, the application of super-resolution reconstruction techniques, particularly those reliant on convolutional neural networks (CNNs), enables the utilization of deep learning's robust feature extraction capabilities to generate high-resolution images from their low-resolution counterparts.

### 3.1 Feature extraction

The feature extraction layer uses convolution operations to extract primary features from the input low-resolution image $I_{LR}$. The main purpose of this layer is to capture the basic information in the image, such as edges, angles, and textures. The feature extraction process is shown in Equation 1.

$$F = f(I_{LR}; \Theta_f) \#(1)$$

Where $f(\cdot)$ represents the convolution operation and $\Theta_f$ is the weight of the convolutional layer. Typically, this layer uses multiple convolution kernels to extract different features, and may include batch normalization and ReLU activation functions to enhance the model's nonlinearity and generalization capabilities. The output $F$ of a convolutional layer is a set of feature maps that provide a wealth of information for subsequent layers. $f(\cdot)$ indicates the convolution operation, which is calculated as shown in Equation 2.

$$f = \sigma(W_f * I_{LR} + b_f) \#(2)$$

Where $*$ is the convolution operation, $W_f$ is the convolution kernel weight, $b_f$ is the bias term, and $\sigma(\cdot)$ is the activation function. $\Theta_f = \{W_f, b_f\}$ is a parameter of the feature extraction layer.

Typically, this layer uses multiple convolution kernels to extract different features and includes batch normalization and ReLU activation functions to enhance the model's nonlinearity and generalization capabilities. The batch normalization process is shown in Equation 3.

$$BN(F) = \gamma \left( \frac{F - \mu_B}{\sqrt{\sigma_B^2 + \epsilon}} \right) + \beta \#(3)$$

Where $\mu_B$ and $\sigma_B^2$ are the mean and variance of the minibatch, $\gamma$ and $\beta$ are trainable parameters, and $\epsilon$ is a very small number in case of dividing by zero. Following Figure 1 shows the general framework of our proposed convolutional model.

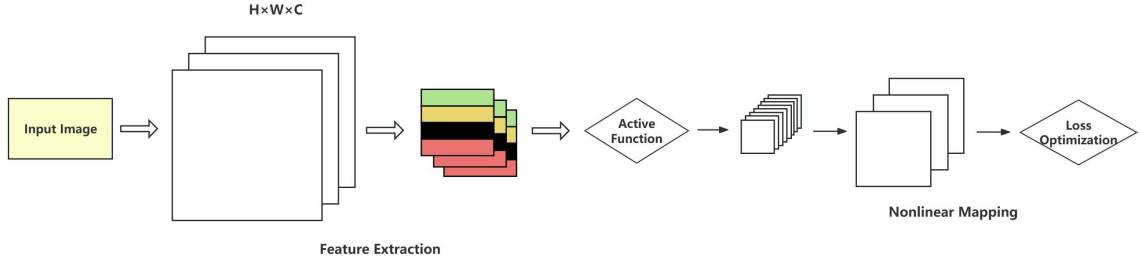

Figure 1. Convolutional Neural Network Framework.

### 3.2 Nonlinear mapping

After feature extraction, feature $F$ is passed to nonlinear mapping layers, which are responsible for transforming these features into a higher layer of feature space $F'$, thus further preparing the reconstruction of high-resolution images. The mapping process is shown in Equation 4.

$$F' = g(F, \Theta_g) \#(4)$$

Where $g(\cdot)$ is a composite function of a series of convolutional layers, typically including multiple convolutional layers, a nonlinear activation function, and, possibly, batch normalization. $\Theta_g$ represents the parameters of these convolutional layers. The specific operation formula for each convolutional layer is expressed as Equation 5.

$$F_{l+1} = \sigma(W_l * F_l + b_l) \#(5)$$

Where $l$ represents the $l$-layer convolution, $W_l$ and $b_l$ are the weights and biases of the layer, respectively. The function $\sigma(\cdot)$ is the activation function. After multilayer convolution and nonlinear activation, features are mapped to a higher-dimensional feature space $F'$. The nonlinear mapping layer processes the feature map $F'$ will be used to generate a high-resolution image. The refactoring process is shown in Equation 6.

$$I_{HR} = \sigma(W_h * F' + b_h) \#(6)$$

The design of these convolutional layers ensures that the output image is dimensionally consistent with the desired high-resolution image.

We utilize the loss function $L(I_{HR}, I_{HR}^{ground\ truth})$ is used to calculate the error between the predicted low resolution image and high resolution image, which is expressed as following Equation 7.

$$L = \frac{1}{N}\sum_{i=1}^{N} ||I_{HR}^{(i)} - I_{HR,\,ground\,truth}^{(i)}||^2 \#(7)$$

The gradient of the loss function with respect to the model parameter $\Theta_f, \Theta_g, \Theta_h$ is then calculated and calculated as $\frac{dL}{d\Theta} = \frac{dL}{dI_{HR}} \cdot \frac{dI_{HR}}{d\Theta}$. Update the parameters using gradient descent and express as Equation 8, where η is the learning rate.

$$\Theta \leftarrow \Theta - \eta \frac{dL}{d\Theta} \#(8)$$

Through the above proposed method, the image super-resolution reconstruction model based on convolutional neural network can effectively recover high-quality high-resolution images from low-resolution images.

## 4 EXPERIMENTS

### 4.1 Experimental setups

In this section, we used the image super-resolution reconstruction method based on convolutional neural network to carry out detailed experiments on the Urban100 dataset. The dataset contains 100 images of complex urban environments and is suitable for evaluating super-resolution techniques. Our model structure includes a feature extraction layer, a nonlinear mapping layer, and a reconstruction layer, using the activation function of ReLU, the optimizer of Adam, the initial learning rate is set to 0.0001, and the loss function is used as mean square error. A total of 50 cycles were trained. With this configuration, the model successfully reconstructed the high-resolution output from the low-resolution input, demonstrating its effectiveness and superiority when working with urban images with rich textures and details. Following Figure 2 shows the used dataset and reconstruction results.

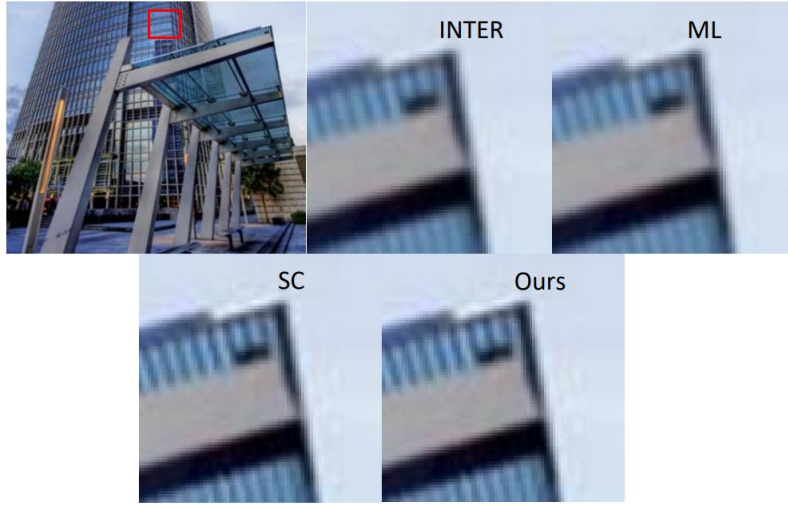

Figure 2. Used Dataset and Simulation Reconstruction Results including INTER, ML, SC and Ours.

## 4.2 Experimental analysis

Peak signal-to-noise ratio (PSNR) is a widely used measure of the quality of image reconstruction or compression, and is particularly useful for comparing the similarity between the restored image and the original image. PSNR is defined by calculating the mean square error (MSE) between the original image and the error image, and using this data, the ratio of the maximum possible power of the signal to the error is calculated, usually expressed in decibels (dB). A higher PSNR value usually indicates a lower error, meaning that the image quality is higher. However, PSNR may sometimes not fully reflect the human eye's perception of image quality, as it is primarily based on numerical errors. Following Figure 3 compares the peak signal-to-noise ratio results.

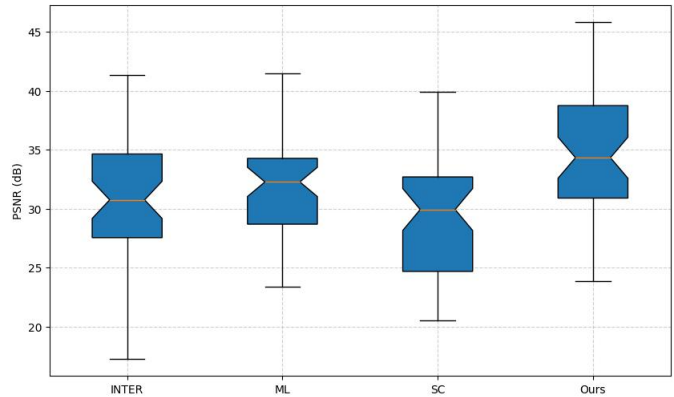

Figure 3. Comparison of PSNR Values Across Different Methods.

Figure 3 shows a comparison of PSNR values for different image super-resolution methods. As shown in Figure 3, "Ours" tend to have a higher PSNR value, indicating that the image reconstruction quality may be better, while "SC" shows a smaller range, which may indicate a poor performance in preserving image detail. This analysis is essential to evaluate the performance of super-resolution algorithms in terms of image quality.

The Structural Similarity Index (SSIM) is a more complex image quality evaluation metric that aims to more accurately simulate the evaluation of image quality by the human visual system. SSIM measures the similarity of two images in terms of visual structure information by considering three dimensions of comparison: brightness, contrast, and structure. SSIM values range from -1 to 1, where 1 means that both images are identical. This method is generally considered to be a more accurate reflection of image quality than PSNR because it takes into account the nature of human visual perception. SSIM is particularly useful for evaluating the effects of processing techniques such as image compression, denoising, and super-resolution. Figure 4 shows the structural similarity comparison results

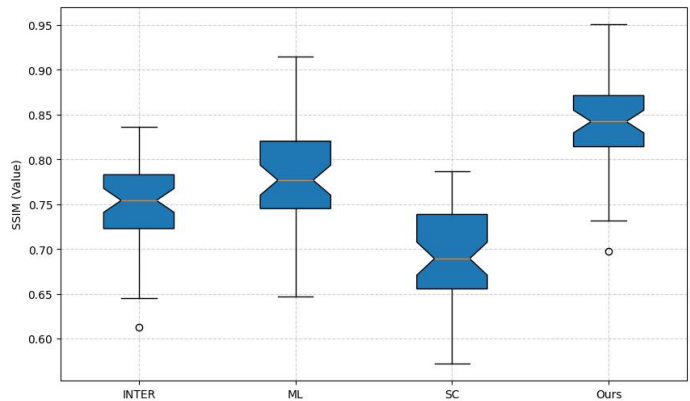

Figure 4. Comparison of SSIM Values Across Different Methods.

Figure 4 shows a comparison of the structural similarity index measurements of different image super-resolution methods. The "Ours" method shows a higher SSIM value, indicating a better performance at maintaining the structural and textural details of the image compared to other methods, while the effectiveness of other methods varies greater.

## 5 CONCLUSION

In conclusion, our study utilizing a Convolutional Neural Network-based mechanism for image super-resolution has demonstrated promising results, particularly evident in the comprehensive evaluations using PSNR and SSIM metrics across various methods such as INTER, ML, SC, and Ours on the Urban100 dataset. The proposed model consistently outperformed other techniques in terms of both PSNR and SSIM, indicating its superior capability to enhance image resolution while maintaining structural integrity and visual quality. These findings underscore the effectiveness of our approach in dealing with complex urban images and highlight the potential of deep learning methods in advancing the field of image super-resolution.